# Selecting Subsets of Source Data for Transfer Learning with Applications in Metal Additive Manufacturing

**Yifan Tang** [1]
Email: yta88@sfu.ca

**M. Rahmani Dehaghani** [1]
Email: mra91@sfu.ca

**Pouyan Sajadi** [1]
Email: sps11@sfu.ca

**G. Gary Wang** [1*]
Email: gary_wang@sfu.ca

[1] Product Design and Optimization Laboratory, Simon Fraser University, Surrey, BC, Canada

## ABSTRACT

Considering data insufficiency in metal additive manufacturing (AM), transfer learning (TL) has been adopted to extract knowledge from source domains (e.g., completed printings) to improve the modeling performance in target domains (e.g., new printings). Current applications use all accessible source data directly in TL with no regard to the similarity between source and target data. This paper proposes a systematic method to find appropriate subsets of source data based on similarities between the source and target datasets for a given set of limited target domain data. Such similarity is characterized by the spatial and model distance metrics. A Pareto frontier-based source data selection method is developed, where the source data located on the Pareto frontier defined by two similarity distance metrics are selected iteratively. The method is integrated into an instance-based TL method (decision tree regression model) and a model-based TL method (fine-tuned artificial neural network). Both models are then tested on several regression tasks in metal AM. Comparison results demonstrate that 1) the source data selection method is general and supports integration with various TL methods and distance metrics, 2) compared with using all source data, the proposed method can find a small subset of source data from the same domain with better TL performance in metal AM regression tasks involving different processes and machines, and 3) when multiple source domains exist, the source data selection method could find the subset from one source domain to obtain comparable or better TL performance than the model constructed using data from all source domains.

## 1 Introduction

Metal additive manufacturing (AM) fabricates parts by depositing metal materials layer by layer with various heat sources, e.g., the laser beam and electric arc. Although metal AM has been adopted in electronics (Pang et al. 2020), automotive (Vasco 2021), aerospace (Blakey-Milner et al. 2021), and other industries, low productivity and unstable quality are two drawbacks that restrict the applications of metal AM. To alleviate the two drawbacks, constructing data-driven models to reveal correlations among processes, structures, and properties has attracted attention in both industry and academia. These models are built based on collected data from experiments or simulations and adopted for process optimization, control, or monitoring to improve the quality of printed parts.

Theoretically, abundant data is required to achieve acceptable accuracy of those data-driven models in metal AM. However, this requirement is hard to satisfy in most cases. In general, experimental data is often limited considering the time-consuming printing process (e.g., a few hours or days per part), while the accuracy of simulation data is usually low due to simplifications (e.g., no geometry deviation) adopted in simulations, which cannot accurately reflect the physical behavior of printing. To reduce the effects of data insufficiency, transfer learning (TL) has been utilized to reuse data collected from completed similar prints to construct the model for the new print (Tang, Rahmani Dehaghani, et al. 2023). In TL, the knowledge in the source domain (e.g., completed prints in metal AM) is extracted and represented as data instances in instance-based TL, feature expressions in

---

feature-based TL, and model parameters or structures in model-based TL. This knowledge is then transferred to the target domain (e.g., new metal AM processes or machines) to build the target model.

TL methods have been applied in metal AM successfully and more accurate target models are obtained than only using limited target data. The instance-based TL and feature-based TL methods have a few applications in geometry prediction (H. Zhang et al. 2021) and relative density estimation (Aboutaleb et al. 2017), while model-based TL methods have rich applications in various metal AM tasks, including geometry prediction (Ren and Wang 2019; Sabbaghi and Huang 2018), defect detection (Brion et al. 2022; Kim et al. 2022; Mehta and Shao 2022; Pandiyan et al. 2022; Senanayaka et al. 2023; Shin et al. 2022; X. Zhu et al. 2023), and melt pool size estimation (Pandita et al. 2022). More details about TL applications in metal AM are discussed in the literature review (Tang, Rahmani Dehaghani, et al. 2023). These applications assume that the source domain is similar to the target domain and all source data are used to construct target models with TL methods directly. In other words, similarity between source and target domains is rarely measured and discussed. Although current applications using all data from one source domain demonstrate the potential of applying TL methods in metal AM, it has been proved in natural language processing (NLP) tasks that a subset of source data could provide better TL performances (Ruder and Plank 2017), which is never explored for metal AM models.

Given one source domain, the simplest strategy is to select the top-$n$ data according to calculated similarities. For example, Aharoni and Goldberg (2020) proposed a domain data selection method, where the similarity between each pair of source and target sentences is quantified by the Cosine distance or the relative similarity of source sentences is inferred by a classification model. The top $n$ pieces of source data with larger similarities (i.e., smaller distances) are then selected for TL. This so-called top-$n$ selection strategy is applied widely, but a single distance metric is hard to perform well among different tasks. To utilize different distance metrics, Ruder and Plank (2017) defined the feature value of each source data as a linear combination of various similarity metrics (e.g., Euclidean distance, Cosine distance, etc.) and diversity metrics (e.g., term distribution and word embeddings). The Bayesian optimization is then adopted to obtain the optimal weights for the linear combination, in order to maximize the performance of TL models. Results on NLP tasks demonstrate that better TL performances are obtained by combining similarity and diversity metrics than only using a single distance metric with all source data. However, the number $n$ is determined manually in the top-$n$ selection strategy, which would deteriorate the TL performance if an improper number is used. Besides, performing Bayesian optimization could be time-consuming when abundant source data are considered. Lin *et al.* (2020) adopted the gradient-boosted decision tree model to infer the relative similarity of each source language from data-dependent features (e.g., dataset size and word overlap) and dataset-independent features (e.g., geographic distance and genetic distance). Based on the ranking of relative similarities of all source languages, the top $n$ source languages provide better TL performances than the baseline only using a single distance metric in tested NLP tasks. However, the accuracy of rankings depends on the decision tree model, which is pre-trained based on TL performances obtained from each pair of source and target languages. When studying tasks where data sizes in both source and target domains are small (e.g., less than 60 in metal AM in Section 3), finding source and target data subsets to pre-train a ranking model is not practical.

When multiple source domains are available, the top-$n$ source domains with larger similarity metrics are selected in current applications. Instead of heuristic selection strategies (e.g., collecting generic data (Mikolov et al. 2019) and a similar field (Karimi et al. 2017)), Dai *et al.* (2019) proposed three measurements to quantify the similarity between source and target data from different aspects, i.e., the percentage of target vocabulary covered by source data, the perplexity of source model on the target domain, and the word vector variance between source and target domains. During tests, those designed similarity measurements reach a consensus on which source domain should be selected to provide a better TL performance. Tian *et al.* (2019) proposed a similarity-based chained TL method for large-scale energy forecasting, where the similarity between each pair of source and target domains (i.e., smart meters) is calculated by Euclidean, Cosign, Manhattan, and Dynamic Time Warping distances. For each target smart meter, the source meter with the largest similarity is selected to construct the target model, which converges faster than training without source data. Instead of selecting one source domain, several source domains with larger similarities are adopted in some applications. Lange *et al.* (2021) proposed a model similarity based on feature vectors (outputs of the last layer) extracted from the source model and the target model, where the target model is trained only with target data. A classification model is then learned to infer which similarity value is related to a positive TL, which means the learning performance on the target domain is improved by TL methods. Given multiple source domains, the source domains classified as ones with positive TL are selected to construct the target model. Lu *et al.* (2023) selected the Dynamic Time Warping distance to infer the similarity between each source and target domain, where the most four similar source domains are selected to construct the multi-source target model. Similar ideas to select one or several similar source domains from multiple source domains are observed widely (Dai et al. 2020; Yuan et al. 2023). Those applications seem to reach a consensus that selecting several similar sources can provide better performance than using a single source domain. However, TL performance obtained by using several similar source domains is not compared with the one obtained by using the optimal subset from one source domain.

There are some other source data selection methods without using distance metrics. For instance, Wang *et al.* (2019) formulated the source data selection as a reinforcement learning task and proposed a minimax game-based data selection method. In the method, the action is the source data selection based on probabilities generated from a selector. The reward is obtained by a discriminator that maximizes the likelihood of distinguishing target and selected source instances. The selector is updated based on rewards with the REINFORCE algorithm (Willia 1992). Although the data selection method provides better TL performances in item recommendation and text matching tasks, the training of the selector is quite expensive, i.e., over $10^6$ steps where a model is constructed in each step. Similar ideas are also observed in task selection for multi-task learning (Guo et al. 2019), weak supervision signals selection for neural information retrieval (K. Zhang et al. 2020), source data selection for domain adaptation (M. Liu et al. 2020), and cross-domain recommendations (Gao et al. 2023). However, the data selection models based on reinforcement learning are trained for certain tasks, which means that the trained model is hard to apply directly to a new task with different source and target domains.

Instead of performing expensive optimization or building data selection models, this paper aims to propose an effective method to find a pseudo-optimal subset from one source domain to get the best TL performances and explore its generalization in metal AM. In specific, this paper aims to achieve the following:

- To develop a source data selection method that does not rely on heuristics and is generally applicable with different distance metrics and TL methods to model metal AM tasks involving various processes and machines.
- To examine the performance of the TL model trained with a well-selected subset of source data in comparison with that trained with the complete source data, and
- To test the conjecture that a smaller TL model using a well-selected subset from a single source domain can have comparable or better prediction accuracy than that trained with all the source data from multiple sources.

The rest of the paper is structured as follows. The applied distance measurements are discussed in Section 2.1, based on which the source data selection method is presented in detail in Section 2.2. To test the generalization capability of the source data selection method, the reproduced TL methods are described in Section 2.3. Several TL tasks in Section 3.2 are designed based on datasets collected from metal AM applications, whose details are discussed in Section 3.1. The TL performances are compared and presented from Section 3.3 to Section 3.5. Limitations and relevant future work of the source data selection method are discussed in Section 4. Finally, Section 5 gives a summary.

## 2 Methods

### 2.1 Distance measurements

The similarity between a source domain and a target domain is generally reflected by distance metrics, where a smaller distance metric refers to a higher similarity. To quantify the relative similarity between each source data and the target domain dataset, two kinds of distance metrics are proposed, i.e., spatial distance and model distance. During the following discussions, the source dataset from the source domain is denoted as $\boldsymbol{D}_s = [\boldsymbol{X}_s, \boldsymbol{Y}_s]$, where $\boldsymbol{X}_s = \{\boldsymbol{x}_s^i, i \in [1, N_s]\}$ and $\boldsymbol{Y}_s = \{y_s^i, i \in [1, N_s]\}$ are source input and output matrices respectively, and $N_s$ is the number of source data. The target dataset from the target domain is $\boldsymbol{D}_t = [\boldsymbol{X}_t, \boldsymbol{Y}_t]$, where $\boldsymbol{X}_t = \{\boldsymbol{x}_t^i, i \in [1, N_t]\}$ and $\boldsymbol{Y}_t = \{y_t^i, i \in [1, N_t]\}$ are target input and output matrices respectively, and $N_t$ is the number of target data.

#### 2.1.1 Spatial distance

A spatial distance quantifies the difference between source and target data in the space. In machine learning, the Euclidean and Cosine distances are the two most common metrics applied to scalars and vectors (Ontañón 2020). The Euclidean distance reflects how far the source data is from the target data, while the Cosine distance indicates the magnitude of the angle between two vectors. Given $\boldsymbol{D}_s$ and $\boldsymbol{D}_t$, the distance metrics between the $i$-th source data $\boldsymbol{d}_s^i = [\boldsymbol{x}_s^i, y_s^i]$ and the target domain are defined as bellow.

(1) Euclidean distance $d_{Euc}^i$

The Euclidean distance $d_{Euc}^i$ between $\boldsymbol{d}_s^i$ and the target domain is defined as the minimum Euclidean distance of the source data $\boldsymbol{d}_s^i$ to all target data, as formulated in Eq. (1),

$$d_{Euc}^i = \min \left( Euclidean(\boldsymbol{d}_s^i, \boldsymbol{d}_t^j) \right), j \in [1, N_t] \tag{1}$$

where $\boldsymbol{d}_t^j = [\boldsymbol{x}_t^j, y_t^j]$ is the $j$-th target data, and $Euclidean(\cdot, \cdot)$ calculates the Euclidean distance between two data. The smaller the $d_{Euc}^i$ value, the higher the similarity between $\boldsymbol{d}_s^i$ and the target data is anticipated.

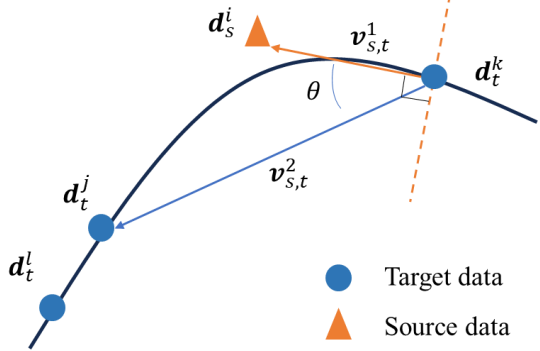

Figure 1 Example of cosine distance in a two-dimensional space

(2) Cosine distance $d_{cos}^i$

The Cosine distance $d_{cos}^i$ between $\boldsymbol{d}_s^i$ and the target domain is formulated as Eq. (2),

$$d_{cos}^i = 1 - \frac{\boldsymbol{v}_{s,t}^1 \cdot \boldsymbol{v}_{s,t}^2}{\|\boldsymbol{v}_{s,t}^1\| \cdot \|\boldsymbol{v}_{s,t}^2\|} \tag{2}$$

where $\boldsymbol{v}_{s,t}^1 = \boldsymbol{d}_s^i - \boldsymbol{d}_t^k$ and $\boldsymbol{v}_{s,t}^2 = \boldsymbol{d}_t^j - \boldsymbol{d}_t^k$ are two vectors defined as shown in Figure 1. The target data $\boldsymbol{d}_t^k$ is the one with the minimum Euclidean distance to $\boldsymbol{d}_s^i$. The target data $\boldsymbol{d}_t^j$ and the source data $\boldsymbol{d}_s^i$ are located on the same side of the dashed line, which is perpendicular to the vector $\boldsymbol{v}_{s,t}^1$. Among all target data on the same side, $\boldsymbol{d}_t^j$ has the minimum Euclidean distance to $\boldsymbol{d}_t^k$. If no source data $\boldsymbol{d}_t^j$ exists, the cosine distance is set as value 1. From the perspective of the response surface between $\boldsymbol{d}_t^j$ and $\boldsymbol{d}_t^k$, the defined Cosine distance is used to reflect the relative amount of information about the target data $\boldsymbol{d}_t^j$ contained by the source data $\boldsymbol{d}_s^i$. A smaller cosine distance $d_{cos}^i$ indicates that more information of $\boldsymbol{d}_t^j$ is presented by $\boldsymbol{d}_s^i$, which refers to a larger similarity.

### 2.1.2 Model distance

Different from calculating distances from source and target data directly, the model distances are defined based on source and target models. The source model is trained only with source data, while the target model is trained only with target data and without any TL method. In this section, the performance distance and the feature distance are defined according to the model perplexity (Dai et al. 2019) and the model similarity (Lange et al. 2021) respectively.

(1) Performance distance $d_{per}^i$

The model perplexity was proposed for language models, where the source model is trained on the source data and the perplexity is defined based on the prediction performance of the source model on the target domain (Dai et al. 2019). Based on this idea, the performance distance $d_{per}^i$ of the source data $\boldsymbol{d}_s^i$ is defined as the prediction accuracy change caused by excluding $\boldsymbol{d}_s^i$ when training the source model. The formulation is shown as,

$$d_{per}^i = error(f_{base}^s(\boldsymbol{X}_t), \boldsymbol{Y}_t) - error(f_{new-i}^s(\boldsymbol{X}_t), \boldsymbol{Y}_t) \tag{3}$$

where $f_{base}^s(\cdot)$ is the baseline source model learned from the source dataset $\boldsymbol{D}_s$. $f_{new-i}^s(\cdot)$ is a source model constructed based on a truncated source dataset $\boldsymbol{D}_s^{-i}$, which is defined as the updated source dataset by excluding $\boldsymbol{d}_s^i$, i.e., $\boldsymbol{D}_s^{-i} = \boldsymbol{D}_s - \boldsymbol{d}_s^i$. The function $error(\cdot)$ is a function to calculate prediction errors between actual and predicted target outputs. Generally, the smaller the prediction error, the more accurate the source model is on the target domain. If $d_{per}^i$ is negative, the prediction accuracy of the source model on the target domain decreases when training without $\boldsymbol{d}_s^i$, which means the source data $\boldsymbol{d}_s^i$ is important to the target domain; otherwise, $\boldsymbol{d}_s^i$ is not important. Therefore, the smaller the performance distance $d_{per}^i$, the larger the similarity between $\boldsymbol{d}_s^i$ and the target domain.

(2) Feature distance $d_{fea}^i$

The model similarity was proposed to learn a mapping from the source feature to the target feature, where the difference between the mapping matrix and the identity matrix is used to quantify the dissimilarity between the source domain and the target domain (Lange et al. 2021). The smaller the matrix difference, the larger the similarity. Based on the definition of model similarity, the feature distance $d_{fea}^i$ is defined as the change of feature differences when training the source model with or without the source data $\boldsymbol{d}_s^i$.

The feature matrix in (Lange et al. 2021) is defined as the output matrix from the last layer of the baseline source model $f_{base}^s(\cdot)$ and the baseline target model $f_{base}^t(\cdot)$. The inputs of both models are the same target testing dataset. As this idea is close to the model perplexity, in this paper the model parameter matrices (e.g., the weight matrix between two successive hidden layers in neural networks) of source and target models are selected as the feature matrices for the feature distance calculation.

The model parameter matrices of $f_{base}^s(\cdot)$ and $f_{base}^t(\cdot)$ are denoted as $\boldsymbol{\theta}_{base}^s$ and $\boldsymbol{\theta}_{base}^t$ respectively. The linear transformation matrix $\boldsymbol{T}_{base}$ is then obtained by solving the following optimization problem with the Procrustes method (Schönemann 1966),

$$\min_{\boldsymbol{T}_{base}} \|\boldsymbol{T}_{base}\boldsymbol{\theta}_{base}^s - \boldsymbol{\theta}_{base}^t\|_F \tag{4}$$

where $\|\cdot\|_F$ is the Frobenius norm. The baseline model dissimilarity $diff_{base}$ is calculated as the difference between $\boldsymbol{T}_{base}$ and the identity matrix $I$, i.e., $diff_{base} = \|\boldsymbol{T}_{base} - \boldsymbol{I}\|$. If the difference is zero, the source model and the target model are identical, indicating the largest similarity.

For each source data $\boldsymbol{d}_s^i$, a new source model $f_{new-i}^s(\cdot)$ with parameter matrices $\boldsymbol{\theta}_{new-i}^s$ is constructed based on the truncated source dataset $\boldsymbol{D}_s^{-i}$, similar to that for defining the performance distance. By solving Eq. (4) with $\boldsymbol{\theta}_{new-i}^s$ and $\boldsymbol{\theta}_{base}^t$, the new linear transformation matrix $\boldsymbol{T}_{new}^i$ is obtained to calculate the new difference as $diff_{new}^i = \|\boldsymbol{T}_{new}^i - \boldsymbol{I}\|$. The feature distance $d_{fea}^i$ of the source data $\boldsymbol{d}_s^i$ is then calculated as $d_{fea}^i = diff_{base} - diff_{new}^i$. If the feature distance is negative, the model dissimilarity increases when training the source model without the source data $\boldsymbol{d}_s^i$. In other words, a small feature distance $d_{fea}^i$ refers to a large similarity of the source data $\boldsymbol{d}_s^i$ to the target domain.

## 2.2 Source data selection method

Compared with using a single distance metric, combining several metrics could capture complementary information (Ruder and Plank 2017). Intuitively, one source data with the largest similarity to the target domain should have the smallest value in each distance metric. However, one source data might have different relative similarities under different metrics, meaning that different distance metrics could conflict with each other. For instance, one source data with a small Cosine distance does not guarantee a small performance distance, as shown in Figure 2, where both Cosine and performance distances are normalized to $[0,1]$. This phenomenon is similar to multi-objective optimization problems, where conflicting multiple objectives prevent obtaining minimum/maximum values of all objectives simultaneously. To find the pseudo-optimal subset, a source data selection method is proposed based on the Pareto frontier by considering different distance metrics.

Figure 2 depicts the source data selection process, regarded as a search method in this paper. For example, the Cosine distance and the performance distance of each source data $\boldsymbol{d}_s^i$ are considered. At the first search step, the subset $\boldsymbol{D}_s^{sel-1}$ on the Pareto frontier of the source dataset $\boldsymbol{D}_s$ is selected and combined with target data to train the target model. In the second search step, the subset $\boldsymbol{D}_s^{sel-2}$ on the Pareto frontier of the updated source dataset $\boldsymbol{D}_s - \boldsymbol{D}_s^{sel-1}$ is selected and combined with $\boldsymbol{D}_s^{sel-1}$ and target data to train the target model. This process continues until finding the pseudo-optimal subset from the source data. Generally, the selected source data at the $k$-th search step is the combination of all subsets located on the former $k$ Pareto frontiers, i.e., $\boldsymbol{D}_s^{sel-1} + \cdots + \boldsymbol{D}_s^{sel-k}$.

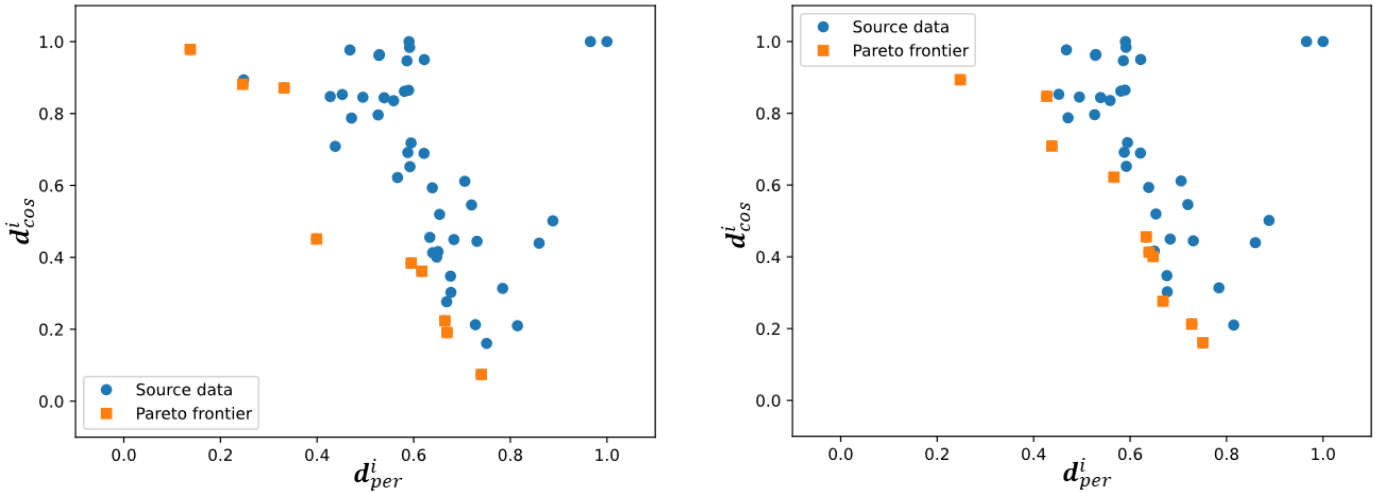

(a) 1$^{\text{st}}$ search step with $\boldsymbol{D}_s$        (b) 2$^{\text{nd}}$ search step with $\boldsymbol{D}_s - \boldsymbol{D}_s^{sel-1}$

Figure 2 Illustration of the source data selection process

Based on the idea, the flowchart of the proposed source data selection method is shown in Figure 3, where the local search and the exhaustive search are considered. The high-level steps are discussed below, and some detailed settings used in this paper are presented in Section 3.2.

*Step 1*: Given the target dataset $\boldsymbol{D}_t$, the baseline target model is constructed without using TL to obtain its baseline prediction error $\sigma_{baseline}$ in the target domain. Based on selected distance metrics, the distances of each source data $\boldsymbol{d}_s^i$ to the target domain is calculated as discussed in Section 2.1. The initial search

*Step 2*: When $k = 1$, $\boldsymbol{D}_s^{sel-1}$ is selected based on the Pareto frontier of $\boldsymbol{D}_s$. When $k \geq 2$, the subset located on the Pareto frontier of the source dataset $\boldsymbol{D}_s - \boldsymbol{D}_s^{sel-1} - \cdots - \boldsymbol{D}_s^{sel-(k-1)}$ is selected and denoted as $\boldsymbol{D}_s^{sel-k}$. Combining subsets $\boldsymbol{D}_s^{sel-1} + \cdots + \boldsymbol{D}_s^{sel-k}$ and target dataset $\boldsymbol{D}_t$, the target model construction is repeated by $N_{run}$ runs to reduce the effects of randomness on modeling. The statistical prediction error $\sigma_k$ of all trained target models on the target domain is used to reflect the overall performance of the selected source data $\boldsymbol{D}_s^{sel-1} + \cdots + \boldsymbol{D}_s^{sel-k}$.

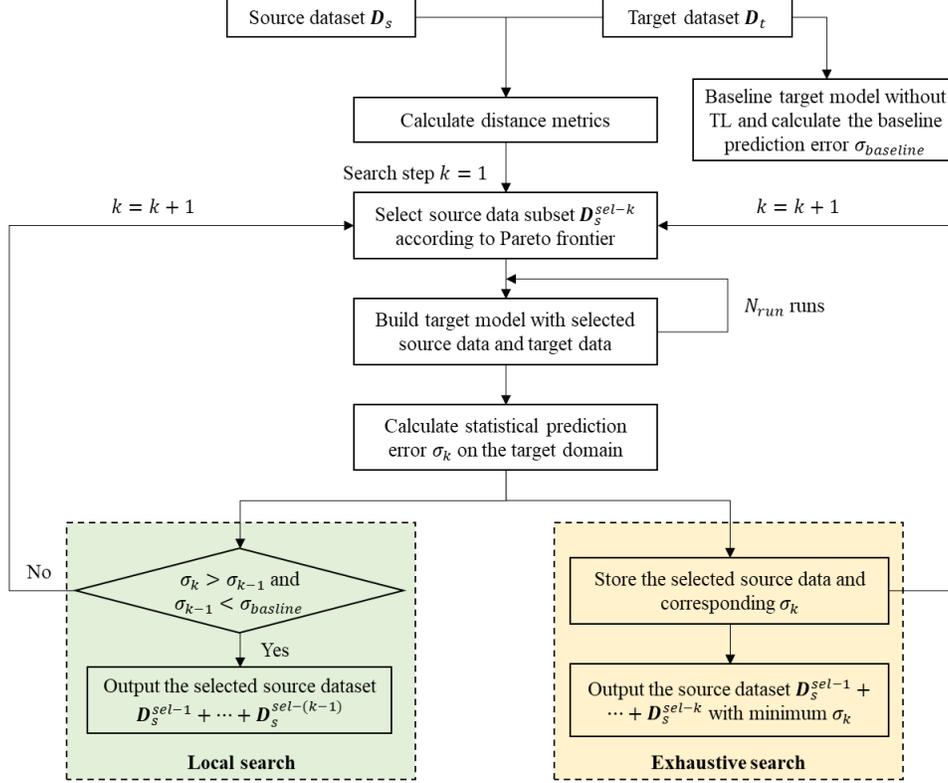

Figure 3 Flowchart of the Pareto frontier-based source data selection method

*Step 3*: This step considers two search pipelines for different purposes.

*Step 3-1 (Local search)*: When the computation resource is limited, the local search is used to find the smallest subset from the source data. The stopping criteria $\sigma_k > \sigma_{k-1}$ and $\sigma_{k-1} < \sigma_{baseline}$ are checked. If they are satisfied, i.e., the prediction error at the current step is larger than the one at the previous step where the prediction error is smaller than the baseline error, the local search terminates. The smallest subset in the source data is defined as $\boldsymbol{D}_s^{sel-1} + \cdots + \boldsymbol{D}_s^{sel-(k-1)}$, which is denoted as the local optimal subset in this paper. Otherwise, the search step is updated as $k = k + 1$ and returns to Step 2.

*Step 3-2 (Exhaustive search)*: When multiple computation resources are available, the exhaustive search pipeline is proposed to find the subset from source data to provide the best TL performance. At the current step, the selected data $\boldsymbol{D}_s^{sel-1} + \cdots + \boldsymbol{D}_s^{sel-k}$ and the corresponding prediction error $\sigma_k$ are stored. The search process then returns to Step 2 directly and $k = k + 1$. After searching all source data, the subset with the minimum prediction error $\sigma_k$ is output and defined as the global optimal subset in the source data.

### 2.3    Tested TL methods

The instance-based TL and the model-based TL methods are selected and reproduced to test the capability of the proposed source data selection method. The state-of-the-art multi-source TL framework is reproduced to demonstrate the effectiveness of the proposed source data selection method in multi-source TL tasks. In this section, all codes are implemented with Python language and the Pytorch library.

#### 2.3.1    Instance-based TL

The instance-based TL methods select a subset from the source data and combine the subset with all target data to train the target model directly. Therefore, the input variables and the outputs in both source and target domains should be identical when using the instance-based TL methods. In metal AM, this method has been applied in several tasks, e.g., transferring geometry knowledge of printed line among various operating conditions in aerosol jet printing by the Two-stage TrAdaBoost.R2 method (H. Zhang et al. 2021), and predicting the relative density of parts fabricated by laser-based AM processes with data collected from publications involving different steel powders and machines (Aboutaleb et al. 2017).

In this paper, the Two-stage TrAdaBoost.R2 (Pardoe and Stone 2010) is selected to build an instance-based decision tree regression (I-DTR) model. The core idea is to update the weights of source and target data by the outer loop (global iteration) and the inner loop (boosting iteration). At the outer loop, both weights of source and target data are updated to increase the possibility of relevant source data being selected for modeling. During the inner loop, only the weights of target data are updated iteratively to reduce the risk of overfitting, and a group of weak regressors is constructed based on source and target data sampled with the updated weights. After finishing the inner loop, the prediction of target testing data is the weighted median outputs obtained by constructed weak regressors, whose prediction error is calculated by the cross-validation framework on the whole target dataset. The final output model is then the group of weak regressors with the minimum cross-validation error during all outer loops. In this paper, the I-DTR model is implemented by the Two-stage TrAdaBoost.R2 code (Ren 2018) and the *DecisionTreeRegressor* function in the *Sklearn* library. The max tree depth is set as six. The maximum numbers of outer loops and inner loops are set as five and ten respectively. More details about the model are described in (Tang, Dehaghani, et al. 2023).

#### 2.3.2    Model-based TL

In model-based TL methods, the source knowledge represented by model structures and parameters is reused in the target domain to facilitate the target model construction and training. The widely applied fine-tuning method shares a part of the pre-trained source model structure and parameters in the target model, whose input and output layers are tailored and trained with limited target data. According to our previous literature review (Tang, Rahmani Dehaghani, et al. 2023), numerous applications of the fine-tuning method have been observed in metal AM tasks, such as defect detection (Bisheh et al. 2022; Kitahara and Holm 2018; Mehta and Shao 2022; Pandiyan et al. 2022; Shin et al. 2022) and geometry deviation prediction (Ferreira et al. 2020; Knüttel et al. 2022; Ren et al. 2021; Ren and Wang 2019; Sabbaghi et al. 2018). Those successful applications demonstrate the capability of the fine-tuning method to transfer knowledge among different processes (Mehta and Shao 2022; Pandita et al. 2022), machines (Knüttel et al. 2022; Ren et al. 2021), geometries (Ferreira et al. 2020; Ren and Wang 2019; Sabbaghi et al. 2018), and materials (Pandiyan et al. 2022; Shin et al. 2022) in metal AM.

This section reproduces the artificial neural network (ANN) with the fine-tuning method, which is denoted as FT-ANN in further discussions. Theoretically, an optimal model structure could be found for each task by optimizing model hyperparameters, such as the number of hidden layers and nodes, the activation functions, the learning rate, and so on. This is the topic of "hyperparameter optimization" in deep learning (Bischl et al. 2023), which is not considered in this paper. Instead, an ANN with three hidden layers is used as shown in Figure 4, where $n_{in}$ is the input dimension and $n_{out}$ is the output dimension. The numbers of hidden nodes on the first, second, and third layers are defined as $2n_{in}$, $3n_{in}$, and $2n_{in}$ respectively, resulting in $14n_{in}^2 + (7 + 2n_{out})n_{in} + n_{out}$ parameters to learn in the model structure. To reduce the risk of overfitting when training data is insufficient, a dropout layer with a probability of 0.05 is applied after the second hidden layer. The reproduced ANN structure is applied to both source and target models, whose training processes are described in Table 1. Generally, the target ANN model is initialized as the trained source model and fine-tuned with limited target data and a smaller number of training epochs. In this paper, the numbers of epochs to train the source model and fine-tune the target model are set as 100 and 50 respectively. The Adam optimizer is used to train both models with a learning rate of 0.005 and the mean square error ($MSE$) is the loss.

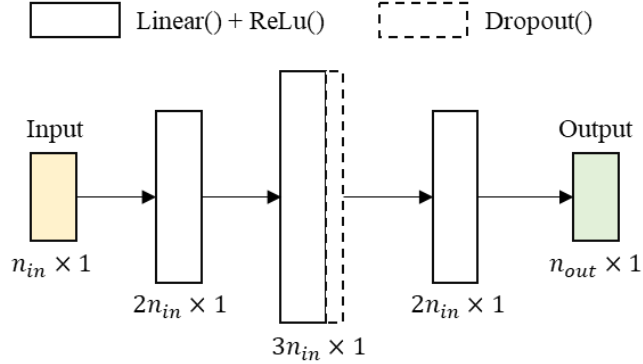

Figure 4 Detailed structure of the artificial neural network for model-based TL

Table 1 Training process of target FT-ANN model

**Input**: source dataset $\boldsymbol{D}_s = [\boldsymbol{X}_s, \boldsymbol{Y}_s]$, target dataset $\boldsymbol{D}_t = [\boldsymbol{X}_t, \boldsymbol{Y}_t]$, the number of epochs to train the source model $epoch_s$, the number of epochs to fine-tune the target model $epoch_t$, the learning rate $lr$, the initialized source model $f_s(\cdot)$

**Output**: the target ANN model $f_t(\cdot)$

1. **For** $epoch$ in [1, $epoch_s$]
2.     Calculate the source prediction $\widehat{\boldsymbol{Y}}_s = f_s(\boldsymbol{X}_s; \boldsymbol{\theta}_s)$, where $\boldsymbol{\theta}_s$ refers to the parameters to learn
3.     Calculate the mean square error $e_s = MSE(\widehat{\boldsymbol{Y}}_s, \boldsymbol{Y}_s)$
4.     Update the source model parameters based on Adam optimizer, i.e., $\boldsymbol{\theta}_s = Adam(\boldsymbol{\theta}_s, e_s, lr)$
7. **End**
8. Initialize the target model based on the trained source model, i.e., $f_t(\cdot; \boldsymbol{\theta}_t), \boldsymbol{\theta}_t = \boldsymbol{\theta}_s$
9. **For** $epoch$ in [1,$epoch_t$]
9.     Calculate the target prediction on the target input, $\widehat{\boldsymbol{Y}}_t = f_t(\boldsymbol{X}_t; \boldsymbol{\theta}_t)$,
10.     Calculate the mean square error $e_t = MSE(\widehat{\boldsymbol{Y}}_t, \boldsymbol{Y}_t)$
11.     Update the target model parameters based on the Adam optimizer, i.e., $\boldsymbol{\theta}_t = Adam(\boldsymbol{\theta}_t, e_t, lr)$
11. **End**

### 2.3.3 Multi-source TL

The multi-source TL methods utilize knowledge extracted from multiple source domains to improve the learning performance on the single target task. One idea is to construct a model based on each pair of source and target domain datasets, and the final target model is the combination of the constructed models. This idea has been successfully applied to object recognition (Yao and Doretto 2010), fault diagnosis (J. Tian et al. 2022), remaining useful life prediction (Ding et al. 2022), and multi-class classification (Kang et al. 2020).

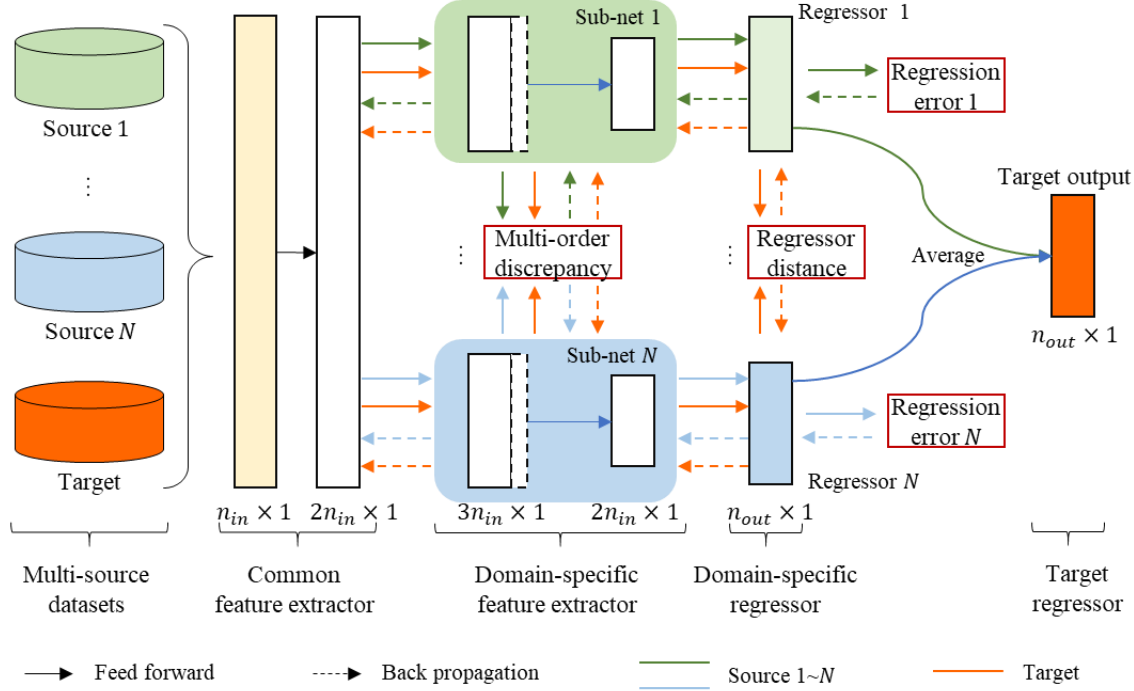

Figure 5 Diagram of the reproduced multi-source ANN model

Based on the framework proposed in the work (Ding et al. 2022), a multi-source ANN (MS-ANN) model as shown in Figure 5 is reproduced based on the model structure in Section 2.3.2. Given $N$ source domains $\boldsymbol{D}_s^i = [\boldsymbol{X}_s^i, \boldsymbol{Y}_s^i], i \in [1, N]$ and one target domain $\boldsymbol{D}_t = [\boldsymbol{X}_t, \boldsymbol{Y}_t]$, the model contains a common feature extractor $F_c$ for all domains, a domain-specific feature extractor $F_{df}^i$ and a domain-specific regressor $F_{dr}^i$ for each source domain $\boldsymbol{D}_s^i$, and a target regressor whose output is the average output from all domain-specific regressors. The numbers of nodes on hidden layers are set as Section 2.3.2, which means a total of $(2 + 12N)n_{in}^2 + (2 + 5N + 2n_{out}N)n_{in} + n_{out}N$ parameters to learn in the multi-source ANN model. Generally, the more source domains are considered, the larger the model structure is. For the training, three components are considered in the loss function, i.e., the multi-order discrepancy, the regressor distance, and the regression errors, which are explained below (Ding et al. 2022).

- *Multi-order discrepancy* $\mathcal{L}_{dis}$. When comparing source and target domains, the maximum mean discrepancy (MMD) calculates the first-order moments of inter-domain distance (Gretton et al. 2012), while the CORAL loss is the second-order statistic between the source and target features (Sun and Saenko 2016). By combining both metrics, the multi-order discrepancy $\mathcal{L}_{dis}$ aims to minimize the domain discrepancy between each pair of the source domain $\boldsymbol{D}_s^i$ and the target domain $\boldsymbol{D}_t$,

$$\mathcal{L}_{dis} = \mathcal{L}_{MMD} + \lambda' \cdot \mathcal{L}_{CORAL} \tag{5}$$

where $\mathcal{L}_{MMD}$ and $\mathcal{L}_{CORAL}$ are functions to calculate MMD and CORAL discrepancies between the source feature $F_{df}^i\left(F_c(\boldsymbol{X}_s^i)\right)$ and the target feature $F_{df}^i(F_c(\boldsymbol{X}_t))$ obtained by the same domain-specific feature extractor $F_{df}^i$. $\lambda'$ is a positive tradeoff coefficient, which is commonly set as 5000 (Ding et al. 2022).

- *Regressor distance* $\mathcal{L}_{reg}$. The regressor distance is used to reduce differences of all domain-specific regressors on the target domain. It is defined as the mean square error (*MSE*) between the outputs of any two different domain-specific regressors on the target input $\boldsymbol{X}_t$, with the following formation.

$$\mathcal{L}_{reg} = \frac{2}{N(N-1) \cdot N_t} \sum_{i=1}^{N-1} \sum_{j=i+1}^{N} MSE\left(F_{dr}^i\left(F_{df}^i(F_c(\boldsymbol{X}_t))\right), F_{dr}^j\left(F_{df}^j(F_c(\boldsymbol{X}_t))\right)\right) \quad (5)$$

- *Regression error $\mathcal{L}_{error}$*. The regression error reflects the performances of the domain-specific regressor on each source domain $\boldsymbol{D}_s^i$. It is adopted to ensure the trained model performs well on each source domain.

$$\mathcal{L}_{error} = MSE\left(F_{dr}^i\left(F_{df}^i\left(F_c(\boldsymbol{X}_s^i)\right)\right), \boldsymbol{Y}_s^i\right) \quad (7)$$

Table 2 Training process of the reproduced MS-ANN model

**Input**: multiple source datasets $\{\boldsymbol{D}_s^i = [\boldsymbol{X}_s^i, \boldsymbol{Y}_s^i], i \in [1, N]\}$, the target dataset $\boldsymbol{D}_t = [\boldsymbol{X}_t, \boldsymbol{Y}_t]$, the maximum number of training epochs $epoch_{max}$, the learning rate $lr$

**Output**: the trained model

1. Calculate the total training steps $step_{max} = N \times epoch_{max}$, initialize the current step as $step = 1$
2. **For** $epoch$ in $[1, epoch_{max}]$
3.     **For** $i = 1, \ldots, N$
4.         batch size $n_b = \min\left(size(\boldsymbol{D}_s^i), size(\boldsymbol{D}_t)\right)$
5.         Randomly sample training data $\boldsymbol{D}_{s'}^i = [\boldsymbol{X}_{s'}^i, \boldsymbol{Y}_{s'}^i]$ and $\boldsymbol{D}_{t'} = [\boldsymbol{X}_{t'}, \boldsymbol{Y}_{t'}]$ from $\boldsymbol{D}_s^i$ and $\boldsymbol{D}_t$
6.         Obtain the source feature by the domain-specific feature extractor $F_{df}^i$, i.e., $F_{df}^i\left(F_c(\boldsymbol{X}_{s'}^i)\right)$
7.         Obtain target features by all domain-specific feature extractors, $F_{df}^1(F_c(\boldsymbol{X}_{t'})), \ldots, F_{df}^N(F_c(\boldsymbol{X}_{t'}))$
8.         Calculate the source prediction by domain-specific regressor $F_{dr}^i$, i.e., $F_{dr}^i\left(F_{df}^i\left(F_c(\boldsymbol{X}_{s'}^i)\right)\right)$
9.         Calculate the outputs of all domain-specific regressors on target data, $F_{dr}^1\left(F_{df}^1(F_c(\boldsymbol{X}_{t'}))\right), \ldots, F_{dr}^N\left(F_{df}^N(F_c(\boldsymbol{X}_{t'}))\right)$
10.        Calculate $\mathcal{L}_{dis}$ based on $F_{df}^i(F_c(\boldsymbol{X}_{s'}^i))$ and $F_{df}^i(F_c(\boldsymbol{X}_{t'}))$ with Eq. (5)
11.        Calculate $\mathcal{L}_{reg}$ based on $F_{dr}^1\left(F_{df}^1(F_c(\boldsymbol{X}_{t'}))\right), \ldots, F_{dr}^N\left(F_{df}^N(F_c(\boldsymbol{X}_{t'}))\right)$ with Eq. (6)
12.        Calculate $\mathcal{L}_{error}$ based on $F_{dr}^i\left(F_{df}^i(F_c(\boldsymbol{X}_{s'}^i))\right)$ and $\boldsymbol{Y}_{s'}^i$ with Eq. (7)
13.        Calculate the total training loss $\mathcal{L}$ with Eq. (8)
14.        Update the model with Adam optimizer and the total training loss
15.        Update the current step as $step = step + 1$
16.     **End**
16. **End**

The final loss $\mathcal{L}$ applied during the training process is defined as,

$$\mathcal{L} = \mathcal{L}_{error} + \gamma \beta_{step} \mathcal{L}_{dis} + \mu \beta_{step} \mathcal{L}_{reg} \quad (8)$$

where $\gamma = 1$ and $\mu = 10$ are the tradeoff coefficients for domain discrepancy and regressor distance (Ding et al. 2022). $\beta_{step}$ is the step-vary tradeoff with the definition $\beta_{step} = 1 + 1/(1 + e^{10*step/step_{max}})$ (Ding et al. 2022). The $step$ refers to the current stage in the training process, and $step_{max}$ is the total number of training steps, defined as the multiplication of the number of source domains with the number of the training epochs. The optimizer Adam with the learning rate 0.005 is adopted to train the model. The number of training epochs is set as the sum of the one to train the source model and the one for fine-tuning in Section 2.3.2, i.e., 150. The detailed training process is described in Table 2. After training, the prediction on the target data is the average output of all domain-specific regressors, i.e., $\widehat{\boldsymbol{Y}}_t = \frac{1}{N} \sum_{i=1}^{N} F_{dr}^i\left(F_{df}^i(F_c(\boldsymbol{X}_t))\right)$.

In general, the reproduced MS-ANN model follows the model framework (e.g., common feature extractor, domain-specific feature extractor, etc.), the loss function, and the training process in (Ding et al. 2022). Compared with work in (Ding et al. 2022), three differences exist in the reproduced model. The components in the reproduced model are selected based on the model-based TL in Section 2.3.2, while convolution neural networks are adopted in the model structure (Ding et al. 2022) for image datasets. The number of training epochs is selected according to the reproduced model-based TL for comparison, while Ref (Ding et al. 2022) uses a large number of training epochs (i.e., 20000). Instead of a fixed batch size applied in (Ding et al. 2022), the batch size $n_b$ in the reproduced model is defined as the minimum number of source and target dataset, as the data sizes in both target and source domains during the source data selection process are small.

## 3    Results

### 3.1    Data preparation

#### 3.1.1    Melt pool width dataset

Two datasets about the melt pool width are considered in this paper. One is about the laser-based directed energy deposition using blown powder (DED-LB/p) published by the Stevens Institute of Technology (Akhavan et al. 2023). Another is about the laser-based directed energy deposition using hot wire (DED-LB/w) collected from University West, Sweden (Dehaghani et al. 2023).

Table 3 summarizes information about the DED-LB/p and DED-LB/w datasets, where three differences are observed. The units of feed material are different, i.e., the mass per unit time is used for the powder in DED-LB/p while the length per unit time is adopted for the wire in DED-LB/w. Meanwhile, the DED-LB/p process only utilizes the laser beam as the heat resource, while the DED-LB/w process uses both the laser beam and electrical power as heat resources. Different from using the laser beam to melt the material in DED-LB/p, the electrical power is used to pre-heat the material before printing in DED-LB/w. Different materials are also used for fabrication, i.e., stainless steel 316L (SS316L) powder in DED-LB/p, and duplex stainless steel 2209 (DSS2209) wire with a diameter of 1.2 mm in DED-LB/w. Given one process parameter setting, images of the melt pool are collected by cameras during printing, based on which time-series melt pool width data is extracted with suitable image processing methods (Akhavan et al. 2023; Dehaghani et al. 2023), as shown in Figure 6. In this paper, the average melt pool width is studied as a result of a given process parameter setting, which is used as input for modeling. Finally, 61 pieces of data are collected from the DED-LB/p dataset and 9 pieces of data are collected from the DED-LB/w dataset. In the following test, the DED-LB/p dataset is defined as the source dataset and the DED-LB/w dataset is the target dataset.

Table 3 Information about DED-LB/p and DED-LB/w datasets

| | DED-LB/p dataset | | DED-LB/w dataset | |
|---|---|---|---|---|
| | | Range | | Range |
| Process parameters | Powder feed rate ($PFR$) | $0.25 - 15$  $g/min$ | Wire feed rate ($WFR$) | $1.8 - 2.2$  $m/min$ |
| | Scanning speed ($SS$) | $10 - 1000$  $mm/min$ | Travel speed ($TS$) | $8 - 12$  $mm/s$ |
| | Laser power ($LP$) | $200 - 500$  $W$ | Laser power ($LP$) | $2800 - 3200$  $W$ |
| | | | Electrical power ($EP$) | $100$  $W$ |
| | | Value | | Value |
| Material | Type | Stainless steel 316L | Type | Duplex stainless steel 2209 |
| | Density ($\rho_{SS316L}$) | $7.98$  $g/cm^3$ (Dongshang 2023) | Density ($\rho_{DS2209}$) | $7.8$  $g/cm^3$  (MatWeb n.d.) |
| Data size | 61 | | 9 | |

Considering different units and numbers of process parameters, it's not reasonable to directly construct TL-

based models with the collected datasets. More specifically, instance-based TL methods cannot support source and target domains with different numbers of input variables (Tang, Rahmani Dehaghani, et al. 2023). To alleviate the problem, some common parameters are defined based on source and target detests, including the material feed rate ($MFR$), the travel speed ($TS$), and the energy density ($ED$). Therefore, the input dimension $n_{in}$ is three and the output dimension $n_{out}$ is one. The formulas of dimensional parameters in both DED-LB/p and DED-LB/w are listed in Table 4. The major difference is that the energy density in DED-LB/w is defined as combining both heat sources (laser beam and electrical power) while only the laser beam is used in DED-LB/p.

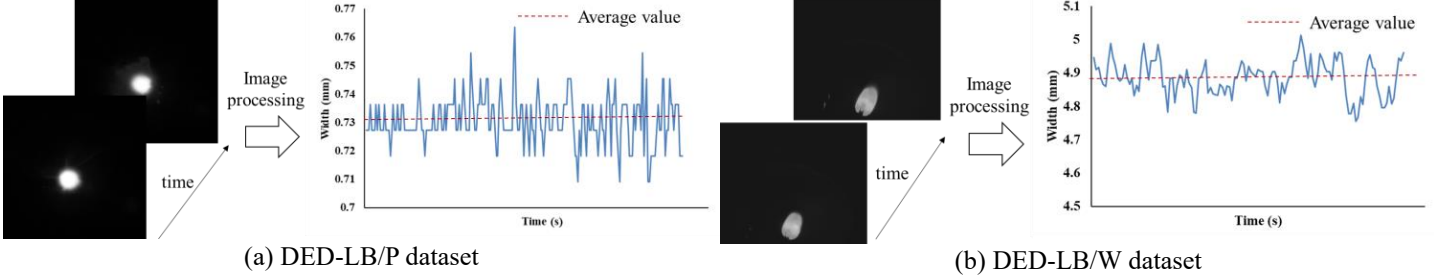

(a) DED-LB/P dataset

(b) DED-LB/W dataset

Figure 6 Time-series melt pool width in DED-LB/P and DED-LB/W datasets

Table 4 Common parameters designed for DED-LB/p and DED-LB/w processes

| Parameter | Unit | Formula | |
|---|---|---|---|
| | | DED-LB/p dataset | DED-LB/w dataset |
| Material feed rate ($MFR$) | $mm^3/s$ | $PFR/\rho_{SS316L} \times \dfrac{1000}{60}$ | $WFR \times 100 \times \pi \times 0.06^2 \times \dfrac{1000}{60}$ |
| Travel speed ($TS$) | $mm/s$ | $SS/60$ | $TS$ |
| Energy density ($ED$) | $J/mm^3$ | $LP/MFR$ | $(LP + EP)/MFR$ |

### 3.1.2  Relative density dataset

Based on multiple publications, Liu *et al.* (2021) proposed an open-source dataset about the relative density in the laser powder bed fusion (LPBF) process, where different printing machines with the same material Ti-6Al-4V are adopted. The ranges of applied process parameters (i.e., laser power, scanning speed, hatching space, and energy density) are summarized in Table 5. The range of each process parameter varies with machines, which is attributed to different properties of machines. After excluding duplicated data, we obtain the data size from each machine, listed in the last column of Table 5. For modeling tasks designed based on this dataset, the input dimension $n_{in}$ is four and the output dimension $n_{out}$ is one. More details about the dataset are described in (S. Liu et al. 2021).

Table 5 Relative density datasets with different printing machines

| Machine | Process parameters | | | | Data size |
|---|---|---|---|---|---|
| | Laser power ($W$) | Speed ($mm/s$) | Hatching space ($mm$) | Energy density ($J/mm^3$) | |
| SLM 125 HL | $50 - 100$ | $300 - 600$ | $0.07 - 0.12$ | $41.7 - 98.8$ | 24 |
| SLM 250 HL | $100 - 375$ | $200 - 1100$ | $0.04 - 0.175$ | $50.62 - 292$ | 32 |
| EOS M270 | $40 - 195$ | $120 - 1560$ | $0.08 - 0.10$ | $17.99 - 288.9$ | 49 |
| SLM | $50 - 120$ | $100 - 1200$ | $0.04 - 0.12$ | $46 - 550$ | 7 |
| Concept Laser M2 | $100 - 200$ | $600 - 1500$ | $0.075 - 0.135$ | $24.7 - 158$ | 22 |
| Concept Laser M3 | $95$ | $90 - 190$ | $0.12 - 0.14$ | $119 - 293$ | 7 |

### 3.2  Test settings

TL-based modeling tasks are designed and summarized in Table 6. The single-source-single-target tasks (i.e.,

Tasks 1-9) are designed to test the generalization of the source data selection method in metal AM tasks with different processes and machines. The multi-source-single-target tasks, i.e., Tasks 10-11, are used to compare the performance of the reproduced multi-source ANN model and the performance of the single-source ANN model with the source data selection method on the same target domain.

Table 6 TL-based modeling tasks

| TL Task | Source domain | Target domain | Dataset |
|---------|---------------|---------------|---------|
| 1 | DED-LB/p process | DED-LB/w process | Melt pool width dataset (Akhavan et al. 2023; Dehaghani et al. 2023) |
| 2 | SLM 250 HL machine | SLM machine | |
| 3 | SLM 125 HL machine | | |
| 4 | EOS M270 machine | | |
| 5 | Concept Laser M2 machine | | |
| 6 | SLM 250 HL machine | Concept Laser M3 machine | Relative density dataset (S. Liu et al. 2021) |
| 7 | SLM 125 HL machine | | |
| 8 | EOS M270 machine | | |
| 9 | Concept Laser M2 machine | | |
| 10 | Any combinations of SLM 125 HL, SLM 250 HL, EOS M270, and Concept Laser M2 machines | SLM machine | |
| 11 | Any combinations of SLM 125 HL, SLM 250 HL, EOS M270, and Concept Laser M2 machines | Concept Laser M3 machine | |

During the following tests, each source and target dataset is scaled to the range [0, 1] for both modeling and distance calculation. The extreme learning machine (ELM) is selected to train the target and source models which are used to calculate both prediction distance and feature distance in Section 2.1.2. The ELM model is a single-hidden layer neural network. Compared with conventional neural networks, the training of ELM is efficient and ELM has a higher possibility to reach the optimal performance (Leung et al. 2019). The error function in the performance distance is defined as $error(\cdot) = 0.5 \times RMSE(\cdot) + 0.5 \times MAE(\cdot)$, where $RMSE(\cdot)$ and $MAE(\cdot)$ are used to calculate the root mean square error (RMSE) and the max absolute error (MAE) between predictions and actual values, respectively. When calculating the feature distance, source and target ELM models share identical parameters for the activation functions. Only the parameters on the output layer are selected to calculate the feature distance for simplification.

In the flowchart of the proposed source data selection method, the number of runs $N_{run}$ to construct a target model based on the given source and target data at each search step is set to 50. Considering the target data is limited in this paper, splitting the target data into training and validation datasets is not reasonable to obtain the performance of the target model. The leave-one-out cross-validation is applied at each run instead. For example, given the target dataset (9 pieces of data) and the selected source dataset at the $k$-th search step in TL Task 1, the selected source dataset and eight pieces of target data are used to construct the target model whose prediction error (i.e., RMSE) is calculated based on the remaining one piece of target data. The overall performance (RMSE) of the target model at each run is then defined as the average RMSE value obtained during cross-validation. The statistical prediction error $\sigma_k$ at the $k$-th search step is the median overall performance of target models constructed in 50 runs. The baseline prediction error $\sigma_{baseline}$ of the target model without TL is calculated under the same process, where 50 runs are repeated and the leave-one-out cross-validation is applied in each run.

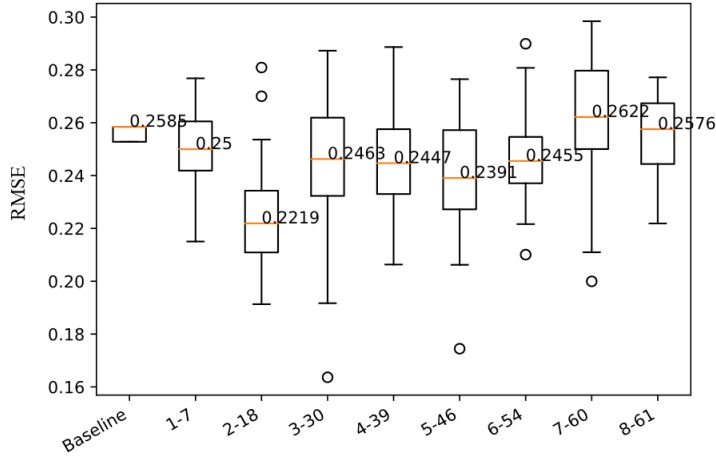

Figure 7 Overall performance of target model among 50 runs at each search step in exhaustive search ("Baseline" refers to the baseline target model. "A-B" refers to the A-th search step with B pieces of source data selected.)

Figure 7 gives an example of the results from the exhaustive search approach (Figure 2), where boxplots of RMSE values obtained by target models at each step (50 runs per step) are depicted. If the local search approach is applied, the search process terminates at the 3rd step according to the criteria in Section 2.2. The source data selected in the former two steps, i.e., $D_s^{sel-1} + D_s^{sel-2}$ with 18 pieces of source data, is output as the local optimal subset. In this example, the global optimal subset is the same as the local optimal subset. Compared with the exhaustive search, the local search terminates earlier which reflects a smaller computational budget. During the following comparisons, the median RMSE value, the amount of data in the optimal subset, and the corresponding search step obtained by both the local and exhaustive searches are selected for comparison.

## 3.3 Generalization of different distance combinations in different TL methods

This section aims to demonstrate the generalization capability of the proposed source data selection method with different distance metrics. To fulfill this purpose, the distance metrics in Section 2.1 are combined to form 11 cases, where six cases are with two distance metrics, four cases are with three distance metrics, and one case is with four distance metrics. All 11 cases are tested with the reproduced I-DTR and FT-ANN models in the TL Task 1, whose performances obtained by local and exhaustive searches are summarized in Table 7 and Table 8.

Compared with the performances of target models trained with all source data in Table 7, the I-DTR model obtains a smaller median RMSE value within three search steps (less than 20 source data) in most cases, where the local search is applied with different distance metrics. The median RMSE values of the target model with all source data are around 0.25, while the median RMSE values of target models with the local optimal subset are less than 0.24 in most cases. Meanwhile, the median RMSE values of I-DTR models using the exhaustive search are smaller than those values obtained with the local search, although the global optimal subset size is larger than that of the local optimal subset in some cases. In other words, the source data selection with both local and exhaustive search approaches can find a subset of source data to construct a more accurate I-DTR model than using all source data.

Results in Table 8 show that the performances of FT-ANN models with the local search are slightly better or comparable to the models using all source data. Different from I-DTR models whose performances are similar between local and exhaustive searches, FR-ANN models with exhaustive search outperform FT-ANN models with local search in terms of the median RMSE values in all cases. Therefore, the proposed source data selection method could also be integrated into FT-ANN for better performance than using all source data.

Table 7 Performance of the I-DTR model in Task 1 with different distance metrics

| Applied distances | Local search | | Exhaustive search | | Result with all source data | |
|---|---|---|---|---|---|---|
| | Median RMSE | # Search step (# source data) | Median RMSE | # Search step (# source data) | Median RMSE | # Search step (# source data) |
| $d_{Euc}^i,\ d_{cos}^i$ | 0.2281 | 1 (3) | 0.2281 | 1 (3) | 0.2592 | 16 (61) |
| $d_{per}^i,\ d_{fea}^i$ | 0.2343 | 2 (6) | 0.2341 | 6 (32) | 0.2633 | 14 (61) |
| $d_{Euc}^i,\ d_{per}^i$ | 0.2359 | 3(16) | 0.2254 | 9 (48) | 0.2555 | 14 (61) |
| $d_{Euc}^i,\ d_{fea}^i$ | 0.2385 | 1 (8) | 0.2342 | 4 (26) | 0.2574 | 10 (61) |
| $d_{cos}^i,\ d_{per}^i$ | 0.2321 | 2 (12) | 0.2321 | 2 (12) | 0.2556 | 13 (61) |
| $d_{cos}^i,\ d_{fea}^i$ | 0.2254 | 1 (10) | 0.2254 | 1 (10) | 0.2483 | 10 (61) |
| $d_{Euc}^i,\ d_{cos}^i,\ d_{per}^i$ | 0.2219 | 2 (18) | 0.2219 | 2 (18) | 0.2576 | 8 (61) |
| $d_{Euc}^i,\ d_{cos}^i,\ d_{fea}^i$ | 0.2256 | 1 (9) | 0.2256 | 1 (9) | 0.2674 | 7 (61) |
| $d_{Euc}^i,\ d_{per}^i,\ d_{fea}^i$ | 0.2491 | 2 (29) | 0.2491 | 2 (29) | 0.2535 | 6 (61) |
| $d_{cos}^i,\ d_{per}^i,\ d_{fea}^i$ | 0.2337 | 1 (16) | 0.2337 | 1 (16) | 0.2556 | 6 (61) |
| $d_{Euc}^i,\ d_{cos}^i,\ d_{per}^i,\ d_{fea}^i$ | 0.2464 | 1 (28) | 0.2464 | 1 (28) | 0.2562 | 4 (61) |

Table 8 Performance of FT-ANN model in Task 1 with different distance metrics

| Applied distances | Local search | | Exhaustive search | | Result with all source data | |
|---|---|---|---|---|---|---|
| | Median RMSE | # Search step (# source data) | Median RMSE | # Search step (# source data) | Median RMSE | # Search step (# source data) |
| $d_{Euc}^i,\ d_{cos}^i$ | 0.1706 | 2 (9) | 0.1697 | 10 (43) | 0.1714 | 16 (61) |
| $d_{per}^i,\ d_{fea}^i$ | 0.1706 | 4 (22) | 0.1669 | 12 (60) | 0.1683 | 13 (61) |
| $d_{Euc}^i,\ d_{per}^i$ | 0.1686 | 7 (37) | 0.1683 | 9 (48) | 0.1699 | 14 (61) |
| $d_{Euc}^i,\ d_{fea}^i$ | 0.1734 | 1 (5) | 0.1683 | 6 (42) | 0.1701 | 12 (61) |
| $d_{cos}^i,\ d_{per}^i$ | 0.1687 | 4 (30) | 0.1679 | 11 (59) | 0.1697 | 13 (61) |
| $d_{cos}^i,\ d_{fea}^i$ | 0.1711 | 1 (5) | 0.1690 | 7 (44) | 0.1721 | 12 (61) |
| $d_{Euc}^i,\ d_{cos}^i,\ d_{per}^i$ | 0.1692 | 2 (18) | 0.1654 | 4 (35) | 0.1715 | 8 (61) |
| $d_{Euc}^i,\ d_{cos}^i,\ d_{fea}^i$ | 0.1694 | 2 (24) | 0.1694 | 2 (24) | 0.1694 | 7 (61) |
| $d_{Euc}^i,\ d_{per}^i,\ d_{fea}^i$ | 0.1675 | 5 (59) | 0.1675 | 5 (59) | 0.1703 | 6 (61) |
| $d_{cos}^i,\ d_{per}^i,\ d_{fea}^i$ | 0.1717 | 1 (6) | 0.1685 | 5 (59) | 0.1716 | 6 (61) |
| $d_{Euc}^i,\ d_{cos}^i,\ d_{per}^i,\ d_{fea}^i$ | 0.1700 | 1 (13) | 0.1669 | 5 (57) | 0.1709 | 6 (61) |

From the above observation, the proposed source data selection method has a generalization capability to consider various distance metrics, while different combinations of distance metrics would provide different TL performance. Meanwhile, the source data selection method is generally applicable and can be integrated into different TL methods to obtain better training performance using a small subset than using all source data.

## 3.4 Generalization in metal AM applications with different machines and processes

The performances of I-DTR and FT-ANN models on Tasks 2-5 (i.e., the target domain is SLM machine) and Tasks 6-9 (i.e., the target domain is Concept Laser M3 machine) are summarized in Table 9 and Table 10, respectively. Generally, both models have better performances with the source data selection method than using all source data directly. When testing those tasks, only the Euclidean distance $d_{Euc}^i$ and the performance distance $d_{per}^i$ are considered in the source data selection method.

Table 9 Performance of I-DTR and FT-ANN models in Tasks 2-5 (SLM machine as the target) with source data selection considering $d_{Euc}^i$ and $d_{per}^i$

| Target model | Source | Local search | | Exhaustive search | | Result with all source data | |
|---|---|---|---|---|---|---|---|
| | | Median RMSE | # Search step (# source data) | Median RMSE | # Search step (# source data) | Median RMSE | # Search step (# source data) |
| I-DTR | SLM 250 HL | 0.0590 | 2 (3) | 0.0590 | 2 (3) | 0.1116 | 10 (31) |
| | SLM 125 HL | 0.0712 | 5 (15) | 0.0712 | 5 (15) | 0.1065 | 8 (24) |
| | EOS M270 | 0.0684 | 5 (26) | 0.0684 | 5 (26) | 0.0908 | 11 (49) |
| | Concept Laser M2 | 0.0723 | 1 (2) | 0.0690 | 5 (10) | 0.0820 | 11 (22) |
| FT-ANN | SLM 250 HL | 0.1031 | 1 (1) | 0.1031 | 1 (1) | 0.1081 | 10 (31) |
| | SLM 125 HL | 0.1030 | 1 (4) | 0.1030 | 1 (4) | 0.1253 | 8 (24) |
| | EOS M270 | 0.1006 | 3 (17) | 0.1006 | 3 (17) | 0.1086 | 11 (49) |
| | Concept Laser M2 | 0.1095 | 1 (2) | 0.1086 | 3 (6) | 0.1134 | 11 (22) |

When the target domain is the SLM machine, the median RMSE values and corresponding search steps of one model are identical between local and exhaustive searches in each task, except for the FT-ANN model in Task 5 (i.e., the source domain is Concept Laser M2 machine). Performances of both I-DTR and FT-ANN models are better with the local optimal subset than with all source data. The number of selected pieces of source data is also less than half of all source data in Tasks 2-5 in most cases. In Tasks 6-9, performances of I-DTR and FT-ANN models show slight differences between local and exhaustive searches, as shown in Table 10. However, similar observations can be made for these results as for Tasks 2-5.

Table 10 Performance of I-DTR and FT-ANN models in Tasks 6-9 (Concept Laser M3 as the target) with source data selection considering $d_{Euc}^i$ and $d_{per}^i$

| Target model | Source | Local search | | Exhaustive search | | Result with all source data | |
|---|---|---|---|---|---|---|---|
| | | Median RMSE | # Search step (# source data) | Median RMSE | # Search step (# source data) | Median RMSE | # Search step (# source data) |
| I-DTR | SLM 250 HL | 0.0066 | 1 (3) | 0.0066 | 1 (3) | 0.0095 | 9 (31) |
| | SLM 125 HL | 0.0064 | 2 (11) | 0.0057 | 6 (23) | 0.0095 | 7 (24) |
| | EOS M270 | 0.0048 | 2 (11) | 0.0036 | 5 (26) | 0.0050 | 11 (49) |
| | Concept Laser M2 | 0.0047 | 1 (5) | 0.0047 | 1 (5) | 0.0067 | 6 (22) |
| FT-ANN | SLM 250 HL | 0.0147 | 2 (9) | 0.0142 | 8 (30) | 0.0153 | 9 (31) |
| | SLM 125 HL | 0.0144 | 1 (4) | 0.0144 | 1 (4) | 0.0164 | 7 (24) |
| | EOS M270 | 0.0130 | 1 (7) | 0.0130 | 1 (7) | 0.0155 | 11 (49) |
| | Concept Laser M2 | 0.0134 | 1 (5) | 0.0134 | 1 (5) | 0.0164 | 6 (22) |

Several conclusions could be summarized from the above tests. When multiple source domains exist for one single target domain, the proposed source data selection method could be used for each pair of source and target domains to improve the modeling performance with fewer source data. Meanwhile, the proposed source data selection method has the general capability to model metal AM processes with TL, where different distances (Section 3.3), as well as machines and processes (Section 3.4), are studied.

### 3.5 Comparison with transfer learning from multiple sources

Table 11 Performance of MS-ANN model on TL Tasks 10-11 with different multiple sources

| Sources | Task 10 (Target: SLM) | | | Task 11 (Target: Concept Laser M3) | | |
|---|---|---|---|---|---|---|
| | 50 epochs | 100 epochs | 150 epochs | 50 epochs | 100 epochs | 150 epochs |
| SLM 250 HL<br>SLM 125 HL<br>EOS M270<br>Concept Laser M2 | 0.1014 | 0.0986 | 0.0991 | 0.0155 | 0.0135 | 0.0134 |
| SLM 250 HL<br>SLM 125 HL<br>EOS M270 | 0.1060 | 0.0990 | 0.0987 | 0.0168 | 0.0137 | 0.0135 |
| SLM 250 HL<br>SLM 125 HL<br>Concept Laser M2 | 0.1056 | 0.0994 | 0.0993 | 0.0170 | 0.0134 | 0.0131 |
| SLM 250 HL<br>EOS M270<br>Concept Laser M2 | 0.1062 | 0.0985 | 0.0997 | 0.0169 | 0.0138 | 0.0132 |
| SLM 125 HL<br>EOS M270<br>Concept Laser M2 | 0.1051 | 0.0982 | 0.0987 | 0.0179 | 0.0140 | 0.0137 |
| SLM 250 HL<br>SLM 125 HL | 0.1156 | 0.0997 | 0.0995 | 0.0195 | 0.0139 | 0.0134 |
| SLM 250 HL<br>EOS M270 | 0.1211 | 0.1000 | 0.0992 | 0.0190 | 0.0142 | 0.0135 |
| SLM 250 HL<br>Concept Laser M2 | 0.1136 | 0.1003 | 0.1006 | 0.0196 | 0.0135 | 0.0130 |
| SLM 125 HL<br>EOS M270 | 0.1270 | 0.0999 | 0.0971 | 0.0190 | 0.0154 | 0.0142 |
| SLM 125 HL<br>Concept Laser M2 | 0.1243 | 0.1001 | 0.1000 | 0.0186 | 0.0144 | 0.0136 |
| EOS M270<br>Concept Laser M2 | 0.1131 | 0.0982 | 0.0979 | 0.0206 | 0.0150 | 0.0139 |

In this section, the designed Tasks 10-11 are tested with the reproduced MS-ANN model. Similar to the above tests, the MS-ANN model construction is repeated by 50 runs, where the leave-one cross-validation is performed in each run. The overall RMSE value of the target model on the target domain in each run is the average RMSE value obtained from cross-validation. One example of the training performance is depicted in Figure 8, where boxplots of RMSE values obtained at 50, 100, and 150 epochs are compared. Generally, the median RMSE value and the size of the boxplot reduce with the epoch number.

To explore the effects of different multiple sources, 11 cases are designed based on the four sources (i.e., SLM 125 HL, SLM 250 HL, EOS M270, and Concept Laser M2 machines) as shown in Table 11. This incluses one case using all four sources, four cases with three sources, and six cases involving two sources. Two observations are found from median RMSE values at different epochs in Table 11. For a given number of epochs, more sources considered would provide a better training performance. For instance, the median RMSE values of MS-ANN at 50 epochs in Task 10 are over 0.110 when using two sources, around 0.106 when using three sources, and around 0.101 when using four sources. Similar behavior also exists in Task 10 with 100 epochs, and Task 11 with 50 and 100 epochs. The reason is that more sources would contain more relevant knowledge, which provides better TL performances (Ding et al. 2022). However, more sources cannot always guarantee a better performance than fewer sources. At 100 epochs in Task 10, MS-ANN with two source domains (i.e., EOS M270 and Concept Laser M2) has a median RMSE value of 0.0982, while MS-ANN with all four sources has a median value of 0.0986. This is because adopting more sources could introduce more irrelevant knowledge, which may deteriorate the training.

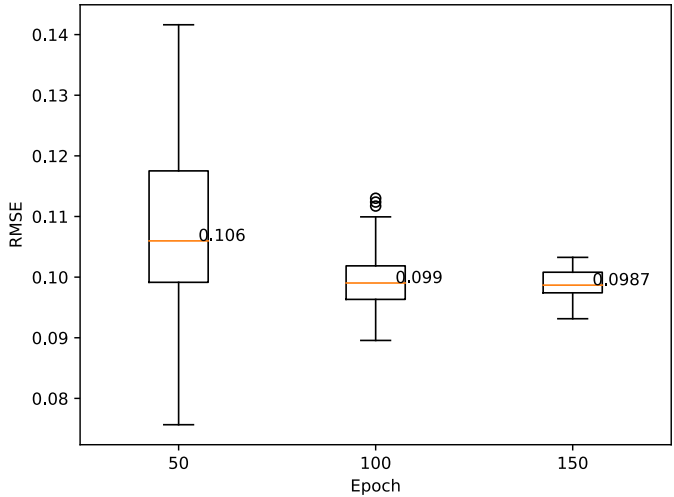

Figure 8 Performances of the MS-ANN model at different training epochs

Table 12 summarizes the ranges of median RMSE values obtained by the MS-ANN at 150 epochs and the FT-ANN using exhaustive search, as well as the corresponding number of model parameters. In most cases, the RMSE values obtained by MS-ANN using multiple source domains are slightly (less than 10%) smaller than those obtained by FT-ANN using a single source domain in both target domains, i.e., SLM and Concept Laser M3 machines. However, the number of parameters in MS-ANN increases a lot with more source domains. Compared to FT-ANN with 261 parameters, MS-ANN has 482 (84.67% larger), 703 (169.3% larger), and 924 (254.02% larger) parameters when using two, three, and four sources, respectively. More importantly, when the target domain is the Concept Laser M3 machine, the FT-ANN with the source domain EOS M270 in Table 10 has a median RMSE value of 0.0130 using the exhaustive search, which is slightly smaller than all median RMSE values (i.e., over 0.0130) of MS-ANN at 150 epochs in Table 11. This observation demonstrates that TL with multiple sources cannot guarantee better performance in all tasks. Instead, the proposed source data selection method could find a small subset from one optional source domain to obtain a better model performance with a smaller neural network.

Table 12 Comparison between FT-ANN and MS-ANN models on the same target domain (input dimension $n_{in} = 4$, output dimension $n_{out} = 1$, number of source domains $N$)

| Models | Range of median RMSE | | Number of model parameters |
|---|---|---|---|
| | SLM | Concept Laser M3 | |
| FT-ANN with a single source ($N = 1$) | [0.1006, 0.1086] | [0.0130, 0.0144] | 261 |
| MS-ANN with two sources ($N = 2$) | [0.0971, 0.1006] | [0.0130, 0.0142] | 482 (+84.67%) |
| MS-ANN with three sources ($N = 3$) | [0.0987, 0.0997] | [0.0131, 0.0137] | 703 (+169.3%) |
| MS-ANN with four sources ($N = 4$) | 0.0991 | 0.0134 | 924 (+254.02%) |

## 4 Discussions

According to results in Section 3, the Pareto frontier-based source data selection method enables integration with different TL methods to model metal AM processes involving various processes and machines. The source data selection method also allows the use of different distance metrics. Meanwhile, the test results reveal that a simple ANN model with the source data selection method using one source domain could outperform the complex multi-source ANN model using several source domains in some cases. Generally, the efficient local search is recommended when the computation resource is limited, while the exhaustive search is suitable to obtain the best performance when multiple computation resources are accessible.

The source data selection method has some limitations, however, in terms of considered distances, TL task categories, and tested TL models. In this paper, two spatial distances (i.e., Euclidean distance and Cosine distance) and two model distances (i.e., feature distance and performance distance) are considered for the metal AM regression tasks. Many other distance metrics are possible, such as term distribution & word embeddings in NLP (Ruder and Plank 2017), dynamic time warping for sequences (Lu et al. 2023), and KL divergence for probability distribution (Ontañón 2020). Although two TL-based models with the source data selection method are tested and outperform the multi-source ANN model in some cases, the neural network reproduced in this paper is simpler than current state-of-the-art models, such as convolutional neural networks (Li et al. 2022), long short term memory model (Van Houdt et al. 2020), and recurrent neural networks (J. Zhu et al. 2022). To explore the generalization capability, more distance metrics, and advanced models could be integrated with the source data selection method and tested on various TL tasks such as time-series forecasting and image classification.

In some works, the distance between each pair of source and target domains is defined as the prediction error of the trained source model on the target domain (Dai et al. 2019), i.e., $error(f_{base}^s(\boldsymbol{X}_t), \boldsymbol{Y}_t)$ in performance distance, or the feature difference between outputs of the same hidden layer in source and target models (Lange et al. 2021), i.e., $diff_{base} = \|\boldsymbol{T}_{base} - \boldsymbol{I}\|$ in feature distance. They are used to infer which source domain or which source domain sets are promising for the transfer learning. In general, the source domain with a smaller distance has a larger similarity to the target domain. Figure 9 compares the base distance $error(f_{base}^s(\boldsymbol{X}_t), \boldsymbol{Y}_t)$ between each source domain (i.e., SLM 125 HL, SLM 250 HL, EOS M270, and Concept Laser M2 machines) and target domain (i.e., SLM, and Concept Laser M3 machines) in the relative density dataset. Based on findings in (Dai et al. 2019; Lange et al. 2021), the source domain with Concept Laser M2 is supposed to provide the best performance for target models, as it has the minimum distances in both target domains. However, this hypothesis is not supported by the test results in this paper, considering the following observations.

- When using all source data, the target I-DTR model for the SLM machine has the smallest median RMSE value 0.0820 with the source domain involving the Concept Laser M2 machine, as shown in Table 9. However, based on all source data from the EOS M270 machine, the target I-DTR model for the Concept Laser M2 machine has the smallest median RMSE value of 0.0050, as shown in Table 10.
- For FT-ANN models in both target domains, applying all source data from SLM 250 HL provides smaller median RMSE values than using all source data from Concept Laser M2, as shown in Table 9 and Table 10.
- Although the target I-DTR model using all source data from Concept Laser M2 has a median RMSE value of 0.0820 in Table 9, the optimal subset from other source domains with larger base distances can provide better modeling performances, i.e., the median RMSE values of 0.0590 and 0.0684 are obtained by using optimal subset from SLM 250 HL and EOS M270 machines respectively.

Therefore, only using a single distance metric is insufficient to compare the transferability of different source domains in all tasks. The above discussions demonstrate that the transferability of one source domain is affected by three factors, i.e., the target domain, the TL method, and the modeling method. Compared with finding the most promising source domain, the proposed Pareto frontier-based source data selection method is more general. The optimal subset from one source domain could be found to obtain better performances than using all data from the most similar source domain. When multiple computation resources are available, the source data selection method could be performed in parallel for each pair of source and target domains. A small model with better or comparable performance could then be obtained instead of training a complex large model using data from all source domains.

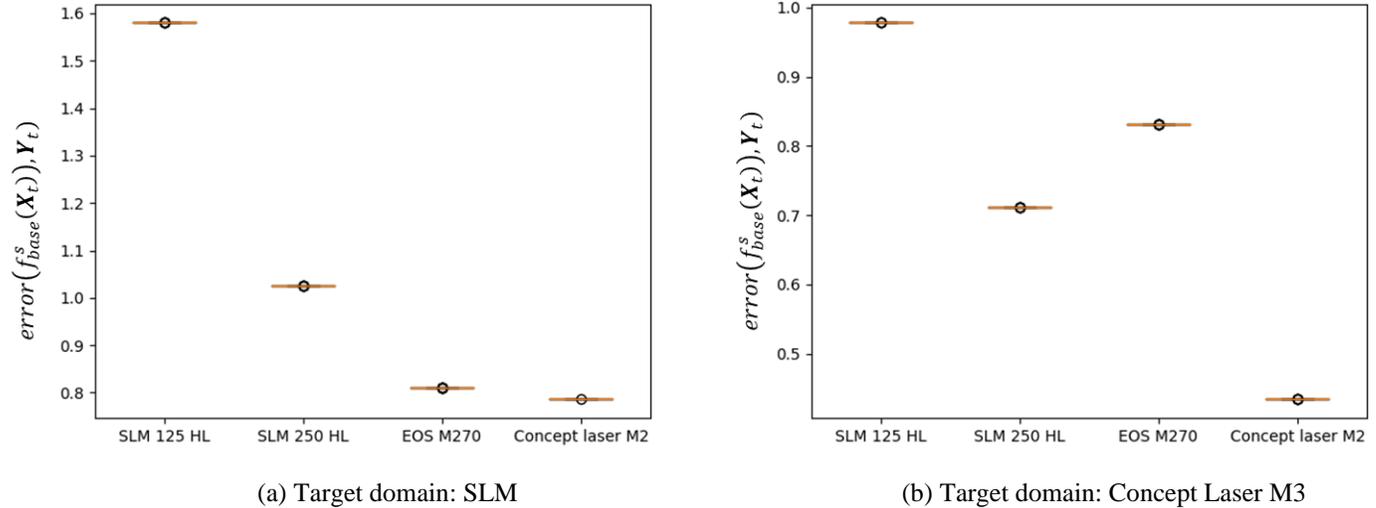

(a) Target domain: SLM

(b) Target domain: Concept Laser M3

Figure 9 Prediction performance of different source domains

# 5 Conclusions

Different from using all data from the source domain in current metal additive manufacturing applications, this paper focuses on finding the pseudo-optimal subset from the accessible source domain to achieve a better transfer learning (TL) performance. To quantify the relative similarity between each target data and the source domain, two spatial distances (i.e., Euclidean and Cosine distances) and two model distances (i.e., performance and feature distances) are adopted. As different distance metrics may conflict with each other, a Pareto frontier-based source data selection method is proposed to iteratively select the source data subset. To test its performance, one decision tree regression TL model with the Two-stage TrAdaBoost.R2 method and one artificial neural network TL model with the fine-tuning methods are reproduced. Meanwhile, one multi-source article neural network is reproduced to use all data from multiple source domains. Based on the melt pool width dataset and the relative density dataset, nine single-source-single-target tasks and two multi-source-single-target tasks are designed for comparison. Several conclusions are summarized from the comparison results:

1) The source data selection method is applicable in various transfer learning methods with different distance metrics,

2) The proposed method can find a small subset of source data (single domain) with better TL performance in metal AM regression tasks involving different processes and machines than using all the data from the source domain, and

3) A model built with the selected subset data from only one source domain has comparable or even higher prediction accuracies with a smaller model size than a model built with data from multiple source domains.

Moreover, the test results demonstrate that the type of source domain, transfer learning method, and modeling method should be considered simultaneously to infer the transferability of one source domain. Instead of finding the most similar source domain, the proposed source data selection method would be a better option when multiple source domains and computation resources are accessible. The proposed source selection method could be tested with more distance metrics, complex models, and transfer learning methods to further study its generalization capability.

**Acknowledgements**

The authors gratefully acknowledge funding from the Natural Sciences and Engineering Research Council (NSERC) of Canada [Grant numbers: RGPIN-2019-06601] and the in-kind support of University West (Dr. Morgan Nilsen and Dr. Fredrik Sikström) under the Eureka! SMART project (S0410) titled "TANDEM: Tools for Adaptive and Intelligent Control of Discrete Manufacturing Processes." Meanwhile, the authors acknowledge the Ph.D. candidate Javid Akhavan at Stevens Institute of Technology, United States, for his help in processing their image datasets.

**Statements and Declarations**

**Competing Interests:** The authors declare that they have no conflicts of interest.

**Data Availability**

The datasets about the melt pool width in DED-LB/p and the relative densities of parts fabricated by different machines are open-accessed. The dataset about the melt pool width in DED-LB/w is available upon reasonable request from the authors.

**Author Contributions**

**Yifan Tang**: Writing – original draft preparation, Validation, Methodology, Investigation, Formal analysis, Conceptualization. **M. Rahmani Dehaghani**: Resources, Methodology, Formal analysis, Conceptualization. **Pouyan Sajadi**: Resources, Methodology, Formal analysis, Conceptualization. **G. Gary Wang**: Writing – review & editing, Supervision, Resources, Project administration, Methodology, Funding acquisition, Conceptualization.